\documentclass[sigconf]{acmart}

\usepackage{booktabs} 

\setcopyright{none}

\usepackage{lipsum}

\acmDOI{}

\acmISBN{}

\acmConference{2nd International Workshop on Embedded and Mobile Deep Learning}{Munich, Germany}{}
\acmYear{June 2018}

\begin{document}
\title{Deploying Deep Neural Networks in the Embedded Space}

\author{Stylianos I. Venieris, Alexandros Kouris and Christos-Savvas Bouganis}
\affiliation{%
  \institution{Dept. of Electrical and Electronic Enigineering, Imperial College London}
}
\email{ {stylianos.venieris10, a.kouris16, christos-savvas.bouganis}@imperial.ac.uk}

\begin{abstract}

Recently, Deep Neural Networks (DNNs) have emerged as the dominant model across various AI applications.
In the era of IoT and mobile systems, the efficient deployment of DNNs on embedded platforms is vital to enable the development of intelligent applications. This paper summarises our recent work on the optimised mapping of DNNs on embedded settings. By covering such diverse topics as DNN-to-accelerator toolflows, high-throughput cascaded classifiers and domain-specific model design, the presented set of works aim to enable the deployment of sophisticated deep learning models on cutting-edge mobile and embedded systems.


\end{abstract}


\maketitle

\vspace{-0.2cm}
\section{Introduction}
\label{sec:intro}

The effective mapping of DNN inference on embedded platforms can offer multiple benefits for mobile AI applications. These \mbox{include:} 1) enabling the processing of high resolution inputs without compromising the user experience due to high-latency access of cloud services, 2) enabling the processing of multiple data sources in high-throughput applications, 3) reducing response time and 4) complying with the power constraints of embedded platforms.

In this context, custom hardware accelerators offer unique opportunities for tailor-made solutions that meet the system-level constraints while providing higher-performance execution and lower power consumption than conventional programmable architectures. However, the inherent complexity of mapping applications on such specialised platforms hinders their adoption from deep learning practitioners. In our recent work, we have tackled a number of critical problems to enhance the accessibility of such platforms. 
The scope of our work ranges from software infrastructure for the automated generation of DNN accelerators to DNN model design with application-specific optimisations. The key categories involve: 1) CNN-to-accelerator toolflows for the automated generation of CNN accelerators, targeting both high-throughput and low-latency settings \cite{venieris2016fccm,venieris2017fpga,Venieris_2017nips}; 2) the exploitation of the resilience of CNNs to low-precision arithmetic to achieve substantial performance gains \cite{kouris2018sysml,kouris2018fpl}; 3) automated synthesis of optimised architectures for emerging multi-CNN applications that employ multiple models for different tasks \cite{venieris2018fpl}; 4) LSTM accelerators that enable deployment in latency-critical applications with limited computation time budget \cite{approxlstm2018arc}; and 5) design of domain-specific DNN models tailored to both the target system's accuracy requirements and compute capabilities \cite{kyrkou2018date}. The rest of the paper presents a high-level view of our work.

\vspace{-0.2cm}
\section{CNN-to-Accelerator Automation}
\label{sec:cnn_to_fpga}
The success of CNNs 
has come with an increase in compute and memory requirements. In this context, FPGAs stand as a promising platform that can meet both the compute requirements and the power constraints of emerging CNN applications. Currently, several obstacles stand as a barrier between deep learning practitioners and FPGAs. From a development perspective, FPGA system development requires expertise in hardware design and familiarity with FPGA toolchains, two skills that typically do not fall within the skillset of deep learning scientists. Moreover, due to the design flexibility of FPGAs, the possible mappings of a CNN on an FPGA lie on a high-dimensional design space that cannot be explored manually. From an application perspective, the diversity of CNN application domains results in a wide spectrum of performance needs. Spanning from the high-throughput needs of multi-sensor systems to latency-critical self-driving cars, the underlying hardware has to be optimised for the particular performance metric of interest. In this context, there is a need for frameworks that abstract the low-level details of FPGAs and automate the generation of FPGA-based CNN accelerators, optimised for the needs of the target \mbox{ application \cite{sv2018csur}}.

 \begin{figure}[t]
	\centering
	\vspace{-0.2cm}
	\includegraphics[trim=4cm 5cm 4cm 2cm, width=0.90\columnwidth,clip]{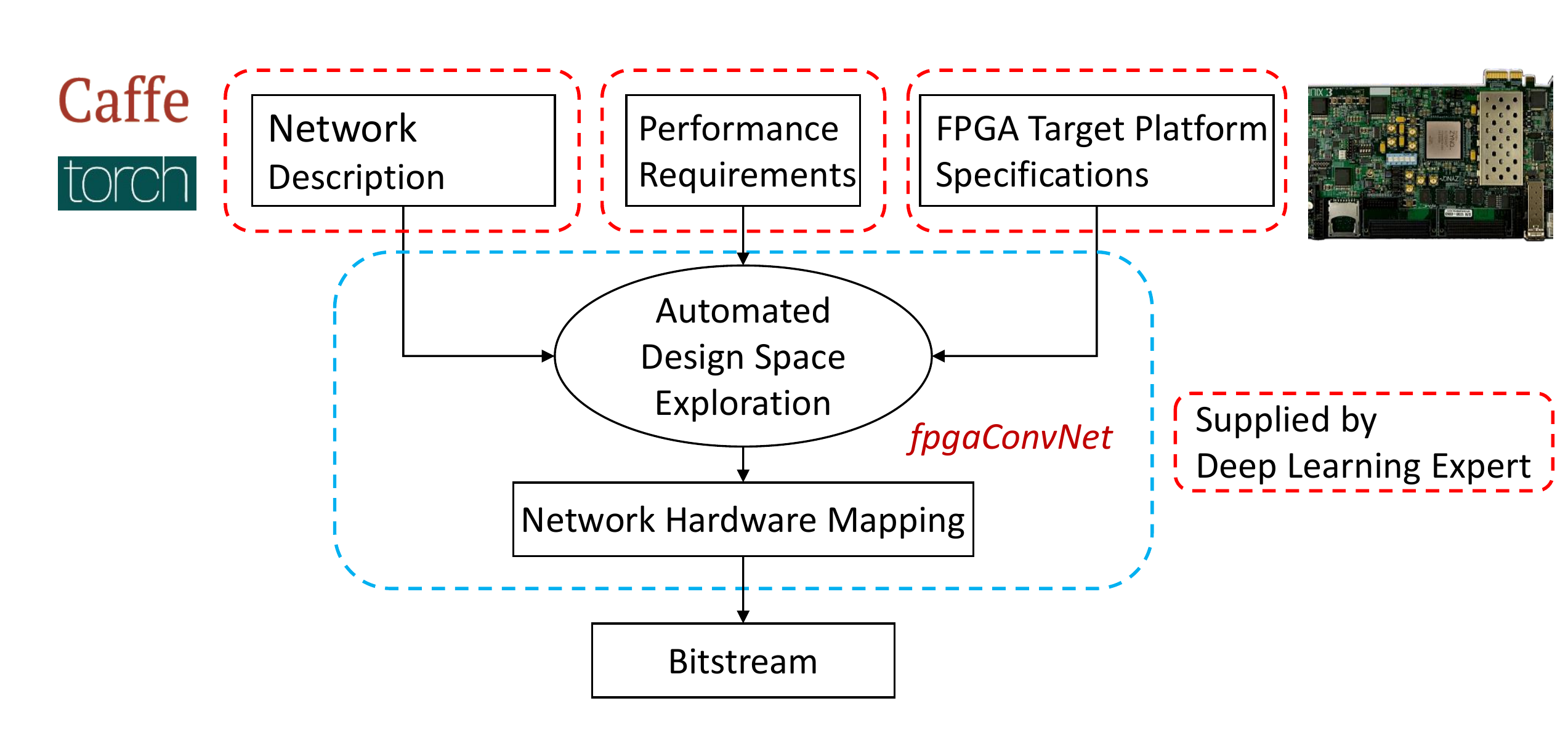}
 	\vspace{-4.5mm}
	\caption{Overview of fpgaConvNet's flow.} 		
	\vspace{-7mm}
	\label{fig:fpgaconvnet}
\end{figure}

\vspace{-0.2cm}
\subsection{The fpgaConvNet Toolflow} 
\vspace{-0.5mm}
\label{sec:fpgaconvnet}
fpgaConvNet is a toolflow whose goal is to automate the mapping of CNNs on FPGAs \cite{venieris2016fccm,venieris2017fpga,Venieris_2017nips,fpgaconvnet2018tnnls}. Starting from a high-level description of a CNN model, fpgaConvNet (Fig. \ref{fig:fpgaconvnet}) considers both the supplied model's workload and the application-level performance needs, including the required throughput and latency, and generates an optimised accelerator for the target FPGA device. At the hardware level, fpgaConvNet employs a highly customisable streaming architectural template which exploits the parallelism both within and across layers and supports large networks by posing no constraints on the model size. To tailor the generated hardware to the CNN-FPGA pair, fpgaConvNet employs an analytical Synchronous Dataflow \mbox{model \cite{Lee_1987}} for capturing both CNN workloads and hardware mappings. This formulation enables the fast exploration of the design space by means of a set of algebraic operations that correspond to a wide range of optimisations and modify the performance-resource cost space of the implementation. To yield the final hardware design, fpgaConvNet casts design space exploration as a mathematical optimisation problem with an objective function that captures the throughput and latency requirements of the target application and automatically generates the resulting hardware design by means of code generation. Overall, in low-power embedded and mobile settings, fgpaConvNet's designs demonstrate performance gains of up to 6.65$\times$ over highly optimised embedded GPU implementations when operating under the same power budget.

\begin{figure}[t] 
	\centering
	\vspace{-0.8cm}
	\includegraphics[trim={2.5cm 9.5cm 3cm 11cm},clip,width=1\linewidth]{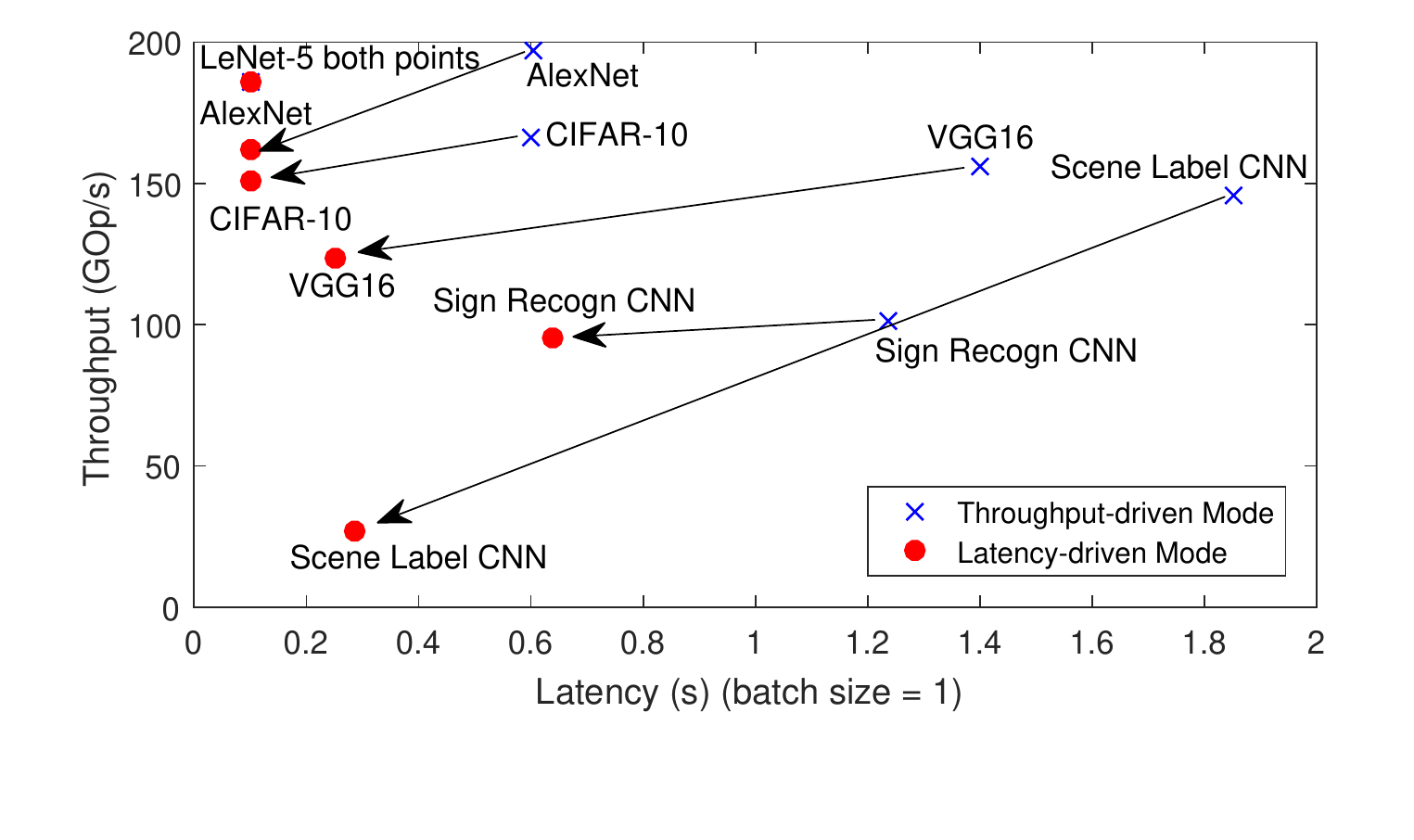}
 	\vspace{-1.5cm}
	\caption{Throughput-driven vs. latency-driven mode.}
	\label{fig:throughput_vs_latency}
	\vspace{-0.4cm}
\end{figure}

\vspace{-0.2cm}
\subsection{A Latency-Driven Methodology}
\vspace{-0.5mm}
\label{sec:latency}
The majority of existing CNN implementations on CPUs \cite{cpu2017throughput}, GPUs \cite{cudnn_2014} and FPGAs \cite{fpdnn2017fccm} are optimised for high throughput. Emerging new AI systems, from autonomous drones \cite{loquercio2018dronet} and cars \cite{kitti2012} to low response-time mobile applications, require the very low-latency execution of CNN inference without the overhead of batch processing. To enable this type of applications, fpgaConvNet 
can place latency at the centre of optimisation and generate latency-optimised hardware designs for the target CNN-FPGA pair \cite{venieris2017fpl}. The latency-driven methodology comprises a run-time configurable architecture that enables the high-performance execution of CNNs without the latency penalty imposed by batching, together with a latency-centric optimiser that guides the design space exploration to previously unreachable low-latency regions (Fig. \ref{fig:throughput_vs_latency}). fpgaConvNet's latency-driven flow delivers up to 73.45$\times$ and 5.61$\times$ improvements in latency over throughput-optimised designs of AlexNet and VGG16 respectively.


\vspace{-0.2cm}
\section{Precision}
\label{sec:precision}
The state-of-the-art accuracy of DNNs in machine vision tasks is achieved at the expense of high computational and memory requirements, that often prohibit their deployment in real-world embedded applications \cite{huang2017speed}. A common strategy to alleviate that cost is the use of reduced precision. The majority of approaches assume the availability of the training set and employ a retraining step to restore the quantised model's accuracy \cite{HanMD15,Guo_2018}. However, in privacy-aware scenarios, such data are not available and hence methods for reducing the precision without retraining are required. Moreover, in applications with low error tolerance, the accuracy loss due to quantisation often prohibits the use of low-precision arithmetic. In this context, there is a need for efficient methods of executing DNNs that combine the gains of reduced precision with negligible accuracy drop.



\vspace{-0.2cm}
\subsection{CascadeCNN's Approach}
\begin{figure}
    \centering
    \vspace{-6mm}
    \includegraphics[trim ={17mm 70mm 35mm 50mm},clip, width=1\columnwidth]{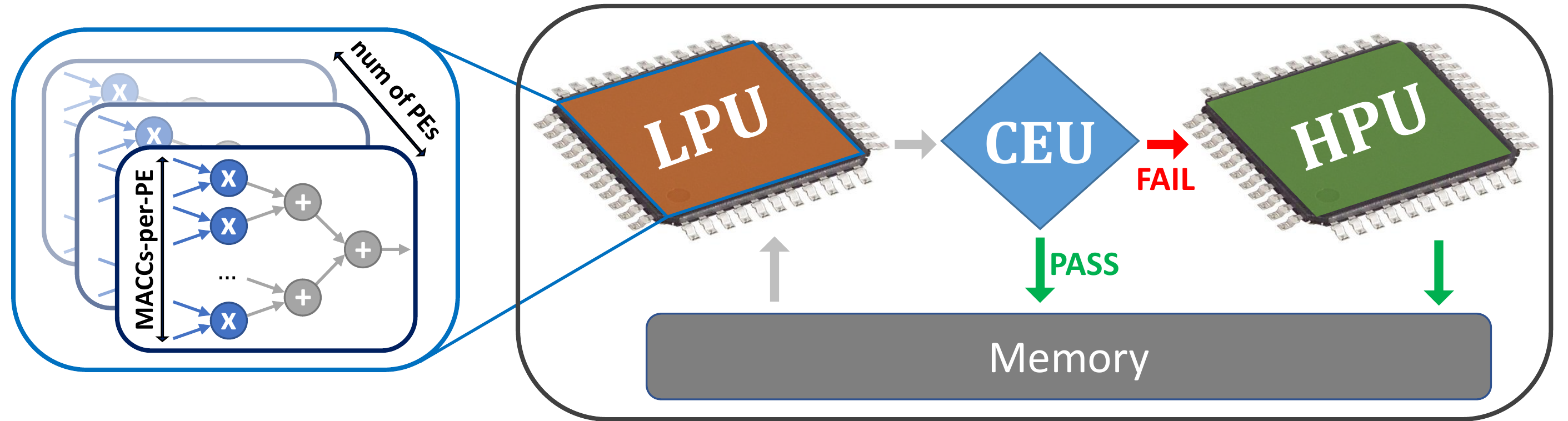}
    \vspace{-6mm}
    \caption{\textit{CascadeCNN}'s high-level architecture. }
    \label{fig:arch}
    \vspace{-5mm}
\end{figure}

\vspace{-0.5mm}
\label{sec:cascade}
\textit{CacsadeCNN} \cite{kouris2018sysml,kouris2018fpl} introduces an automated toolflow for generating a high-throughput cascade of CNN classifiers that pushes the performance of precision-quantised CNNs. Our key observation is that not all inputs of a CNN require the same level of precision in the computation to yield a confident prediction. In this respect, \textit{CascadeCNN} exploits this property and generates a two-stage architecture of precision-quantised models (Fig. \ref{fig:arch}). The first stage consists of an excessively low-precision processing unit that enables rapid classification prediction. The outputs from the first stage are fed to a confidence evaluation unit which estimates the prediction confidence. The samples that are detected as misclassified are recomputed on a high-precision unit to restore the application-level accuracy and comply with the user-specified error tolerance. Overall, \textit{CascadeCNN} considers the error tolerance and the input CNN-FPGA pair to select quantisation scheme, configure the confidence evaluation mechanism and generate the cascaded low- and high-precision processing units. The \textit{CascadeCNN}'s designs demonstrate a performance boost of up to 55\% for VGG16 and 48\% for AlexNet over baseline designs achieving the same accuracy.

\vspace{-0.2cm}
\section{Multi-DNN Systems}
\label{sec:multidnn}
In the construction of complex AI systems, DNN models are used as building blocks of a larger system. 
In this respect, multi-DNN systems have emerged, employing several models, each one trained for a different subtask. In particular, in the emerging field of intelligent autonomous systems, such as drones \cite{Smolyanskiy_2017} and self-driving cars \cite{Chen2015}, the system's perception is largely based on computer vision tasks, such as object detection \cite{Ren_2017} and semantic and instance segmentation \cite{Badrinarayanan_2017, maskrcnn2017cvpr}. Such systems require the concurrent execution of these subtasks and hence the parallel and continuous execution of multiple models.

Nevertheless, deploying multiple models on a target platform poses a number of challenges. From a resource allocation perspective, with each model targeting a different task, the performance constraints, such as required throughput and latency, vary accordingly. Instead of being model-agnostic, this property requires the design of an architecture that captures and reflects the performance requirements of each model. Moreover, in resource-constrained setups, multiple DNNs compete for the same pool of resources and hence resource allocation between models becomes a critical factor. In this respect, the mapping of multiple DNNs is a multidimensional design problem that encompasses both the performance needs of each model and the resource constraints of the target platform.

\vspace{-0.2cm}
\subsection{f-CNN$^\text{x}$: Deploying Multiple CNNs}
\begin{figure}
\vspace{-2mm}
	\centering
	\vspace{-0.8cm}
	\includegraphics[trim={0cm 5cm 11cm 2cm},clip,width=0.5\textwidth]{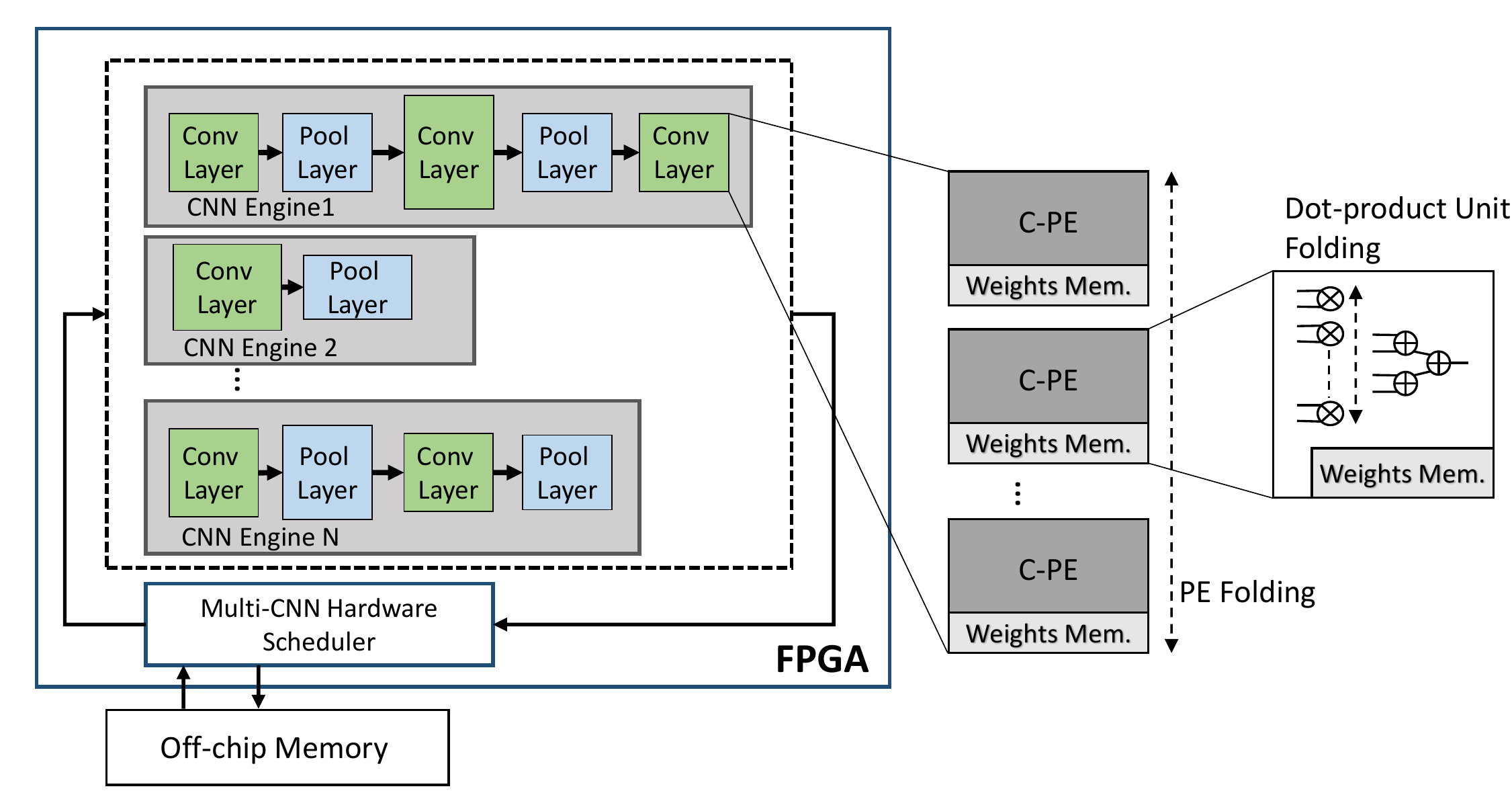}
	\vspace{-0.7cm}
	\caption{Parallel architecture for multiple CNNs.}
	\label{fig:multicnn_arch}
		\vspace{-0.4cm}
\end{figure}

\vspace{-0.5mm}
\label{sec:fcnnx}
f-CNN$^\text{x}$ \cite{venieris2018fpl} is a toolflow which addresses the challenge of mapping multiple CNNs on a target FPGA platform while meeting the required performance for each model. f-CNN$^\text{x}$ exploits the structure of CNN workloads and the fine-grained control over resource allocation of FPGAs to yield latency-optimised designs. From a hardware perspective, the developed toolflow introduces a highly parametrised multi-CNN architecture (Fig. \ref{fig:multicnn_arch}) that allows the fine-grained allocation of resources among CNNs and the deterministic scheduling of competing memory transfers. f-CNN$^\text{x}$ explores a wide range of resource and bandwidth allocations and incorporates the application-level importance of each model by means of multiobjective cost functions to guide the design space exploration to the optimum hardware design. Overall, f-CNN$^\text{x}$ overcomes the limitations of other parallel platforms by yielding up to 6.8$\times$ gains in performance-per-Watt over highly optimised embedded GPU designs in multi-CNN settings. To the best of our knowledge, this work addresses for the first time in the literature the latency-optimised mapping of multiple CNNs.

\section{Computing under Time Constraints}
\label{sec:time_constraints}

Modern intelligent systems such as mobile robots and UAVs that employ DNNs to perceive and interact with their surroundings often operate under time-constrained, latency-critical settings. In such scenarios, the output of a DNN would typically yield an action, such as a manoeuvre to avoid an obstacle. For such decision-making to happen both in real-time and with the best possible outcome, obtaining the most informative output from a DNN given a constraint in computation time is vital. In this respect, the design of computing systems that exploit the runtime-accuracy trade-off of DNNs is necessary to enable the timely operation of such systems.


\vspace{-0.2cm}
\subsection{Approximate FPGA-based LSTMs}
\begin{figure} 
	\centering
	\vspace{-0.9cm}
	\includegraphics[width=0.87\columnwidth, trim={250mm 60mm 230mm 80mm},clip]{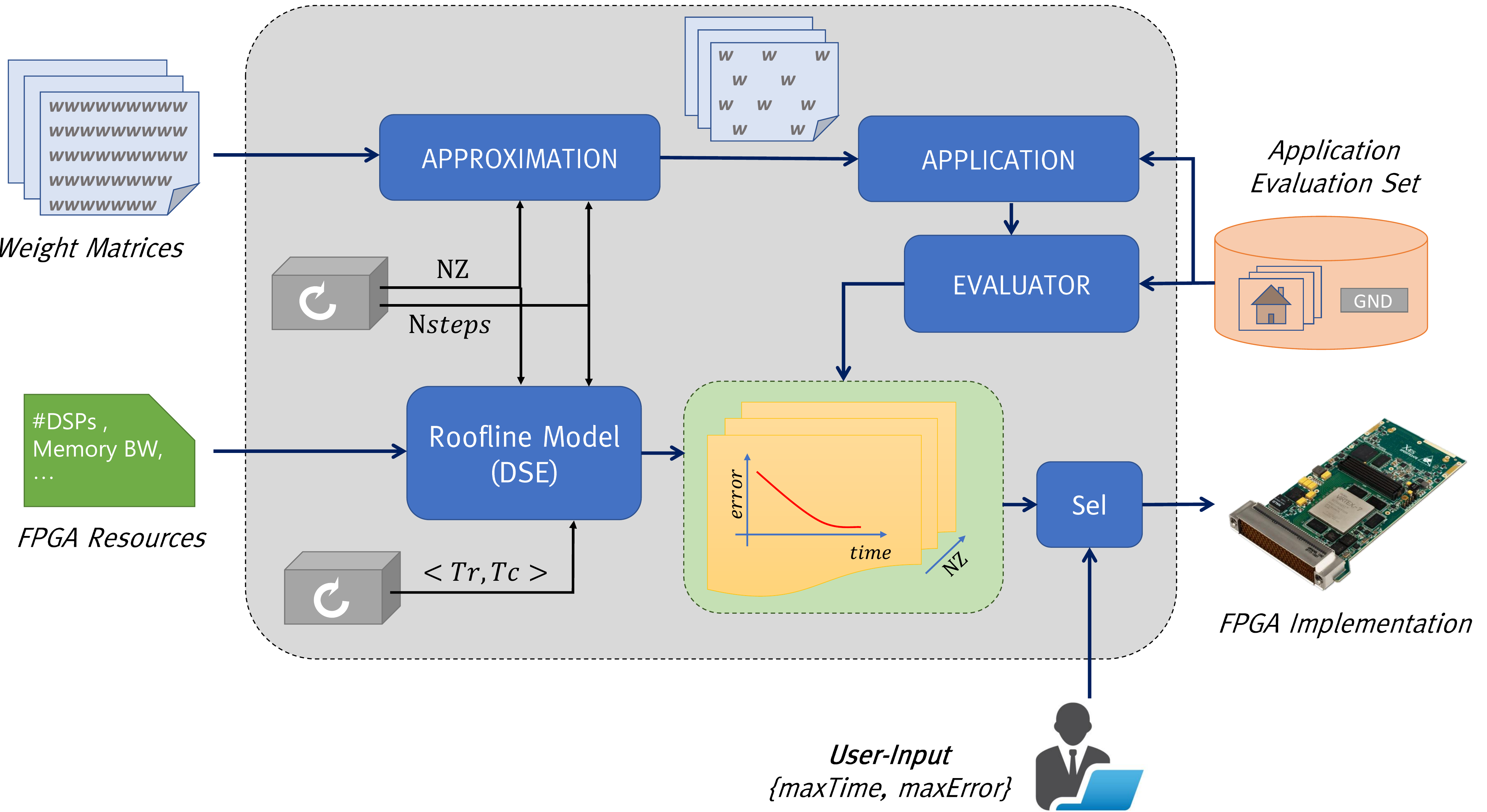}
    \vspace{-0.35cm}
	\caption{Overview of approximate LSTMs design flow.}
	\label{fig:flow}
    \vspace{-0.5cm}
\end{figure}

\vspace{-0.5mm}
\label{sec:approx_lstms}
With a focus on the high-performance deployment of LSTMs under time-constrained settings, \cite{approxlstm2018arc} presents a framework that comprises an approximate computing scheme together with a novel FPGA-based hardware architecture for LSTMs. The proposed framework (Fig. \ref{fig:flow}) employs an iterative approximation method to compress and prune the target LSTM and explore the computation time-accuracy trade-off. Internally, the framework co-optimises the LSTM approximation and the hardware design in order to meet the computation time constraints. By targeting a real-life image captioning application, the designs generated by the developed framework demonstrate 6.5$\times$ less time to achieve the same application-level accuracy over a baseline accelerator, while reaching an average of 25$\times$ higher accuracy under the same computation time constraints.

\vspace{-0.2cm}
\section{Domain-Specific DNN Design}
\label{sec:domain_dnn}
Conventionally, the deep learning community's design methodology for DNN models focuses on maximising the accuracy on the target task, while largely neglecting the implications on the inference-time computational cost. By considering domain-specific properties in order to guide the design of DNNs, more efficient models can be constructed which both meet the required accuracy and lie within the compute capabilities of the target platform.


\vspace{-0.2cm}
\subsection{The DroNet Vehicle Detector}
\vspace{-0.5mm}
\label{sec:dronet}
Drones are emerging as a promising technology for a broad range of applications from domains such as agriculture, security, emergency response and infrastructure monitoring \cite{michael2014collaborative} \cite{nikolic2013}. A prominent application includes the detection of vehicles for emergency response and traffic monitoring. In this scenario, drones operate autonomously and employ CNN models for the detection of vehicles. \cite{kyrkou2018date} presents an end-to-end investigation of different single-shot CNN detectors for drone-based vehicle detection. Starting from the dataset collection and model design down to the deployment on the Odroid-XU4 and Rasperry Pi 3 embedded hardware platforms, this work presents an exploration over the structure of the CNN. To find the CNN that optimises both accuracy and computation cost, a custom metric is employed that, given a model instance, captures both the detection accuracy and the achieved runtime on the target hardware platform. By following this methodology, the resulting detection CNN yields the highest performing balance between detection accuracy and fast execution on the two target embedded platforms.

\vspace{-0.20cm}
\section{Conclusion}
\vspace{-1mm}
The presented set of works focuses on bridging the gap between the deep learning community and the deployment of models in the embedded space. The fpgaConvNet toolflow automates the optimised generation of FPGA-based CNN accelerators targeting both high-throughput and latency-driven applications. In the context of quantised CNNs, \textit{CascadeCNN} alleviates the need for training data availability and enables in this way the use of high-throughput CNN accelerators that employ extremely low-precision arithmetic in privacy-critical applications. By targeting multi-CNN systems, \mbox{f-CNN$^\text{x}$} paves the way in executing multiple CNNs under latency constraints on FPGAs. Moreover, an approximate computing methodology for the deployment of LSTMs in time-constrained applications is presented, enabling latency-critical systems to make informed decisions in real-time. Finally, by considering domain-specific optimisations, a model design methodology has been developed to construct DNNs that are optimised for both the task-level accuracy and compute capabilities of the target platform.


\vspace{-0.2cm}

\bibliographystyle{ACM-Reference-Format}
\bibliography{references}

\end{document}